\documentclass[journal, onecolumn]{IEEEtran}
\usepackage{amsmath,amsfonts}
\usepackage{algorithmic}
\usepackage{algorithm}
\usepackage{array}
\usepackage[caption=false,font=normalsize,labelfont=sf,textfont=sf]{subfig}
\usepackage{textcomp}
\usepackage{stfloats}
\usepackage{url}
\usepackage{verbatim}
\usepackage{graphicx}
\usepackage{cite}
\usepackage{amssymb}
\usepackage{booktabs} 
\usepackage{multirow}
\usepackage{threeparttable}
\usepackage{makecell}
\usepackage{tabularx}
\usepackage{soul}
\usepackage[svgnames, dvipsnames]{xcolor}
\newtheorem{definition}{Definition}

\begin{document}

\title{KAPPA: A Generic Patent Analysis Framework with Keyphrase-Based Document Portraits}

\author{Xin Xia, Yujin Wang, Jun Zhou, Guisheng Zhong, Linning Cai, Chen Zhang

\thanks{Xin Xia, Yujin Wang, Linning Cai and Chen Zhang are with the Department of Industrial Engineering at Tsinghua University, Beijing, China. }
\thanks{Jun Zhou and Guisheng Zhong are with China Intellectual Property Society, Beijing, China. }
\thanks{Corresponding author: Chen Zhang.}}

\markboth{}%
{Xin Xia, Yujin Wang, Jun Zhou, Guisheng Zhong, Linning Cai, Chen Zhang \MakeLowercase{\textit{et al.}}: }


\maketitle
\begin{abstract}
Patent analysis demands document representations that are not only concise but also interpretable by experts. We refer to these representations as portraits. Keyphrases, both present and absent, are ideal candidates for patent portraits due to their brevity, representativeness, and clarity. In this paper, we introduce KAPPA, an integrated framework designed to construct keyphrase-based document portraits and enhance patent analysis. KAPPA operates in two phases: patent portrait construction and portrait-based analysis. For effective portrait construction, we propose a novel semantic-calibrated keyphrase generation paradigm built on pre-trained language models, complemented by a prompt-based hierarchical decoding strategy that leverages the structural features of patents. Comprehensive experiments on benchmark datasets and real-world patents reveal that our keyphrase-based portraits effectively encapsulate domain-specific knowledge and enrich semantic content. Our model demonstrates significant improvements in KP on patents over state-of-the-art baselines.
\end{abstract}
\section{Introduction}
According to the 2024 World Intellectual Property Organization (WIPO) report, global patent applications have increased for the fourth consecutive year, surpassing 3.55 million\footnote{https://www.wipo.int/ipstats/en/}. This growth highlights the importance of patents as a data source for analyzing innovation, and in the meanwhile emphasizes the need for analytical techniques to handle the rising volume of patent data. 
Patent documents are inherently complex, featuring lengthy texts and specialized knowledge, making manual analysis both challenging and time-consuming. To address this, automatically extracting structured representations that contain key information from each patent is essential, which lays the foundation for efficient processing and analysis of patent data. 

 The patent representations should be both concise and interpretable while capturing both salient and latent information within the document. We refer to this as constructing a \textit{patent portrait}. Although some prior works have attempted to build such representations, they often focus on salient information while neglecting latent aspects \cite{tpe, techpat}, or rely on numeric embeddings that lack interpretability \cite{copate}.
Latent information often captures abstract features shared among patents, and its omission can obscure correlations between patents. 
Additionally, numeric embeddings lack interpretability, making them difficult for human experts to comprehend, verify, and refine. This limitation also undermines reliability in scenarios requiring high-quality results.

Keyphrases emerge as promising candidates to construct patent portraits. They serve as critical elements, providing a high-level, concise, and interpretable representation of patent content. Keyphrases offer an accessible overview while maintaining relevance to the document's themes. 
They can be categorized into two types. The first type is present ones which exist verbatim within the document and directly reflect its content. The second type is absent ones which do not appear explicitly in the document but summarize its overarching themes. 

\begin{figure}[h]
  \centering
  \includegraphics[width=\linewidth]{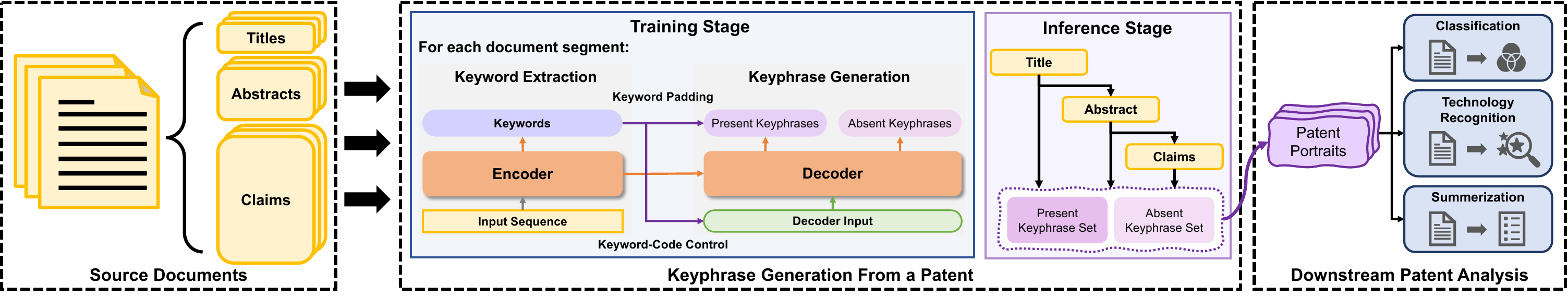}
  \caption{The overall framework of KAPPA.}
  \label{fig:kappa}
\end{figure}

Constructing patent portraits requires generating high-quality keyphrases from patents. To realize that, keyphrase generation (KG) methods need to be adapted to address the distinct characteristics of patent documents, including their structured format and domain-specific language.
Patent documents are semi-structured, comprising multiple levels such as \textit{Title}, \textit{Abstract}, and \textit{Claims} \cite{techpat}. Each level varies in text length and information richness, leading to a distinctive distribution of content. For instance, technical details, often represented by present keyphrases, may span across various levels. In contrast, high-level technologies, captured by absent keyphrases, emerge from the interconnections between these levels. 
This semi-structured format of patents introduces unique challenges for KG. Directly concatenating all levels for KG is impractical due to the length of patent documents. Instead, an effective KG method should efficiently utilize the distributed information across different levels of the document. Moreover, the interconnected nature of the multi-level structure should be used to predict absent keyphrases that summarize high-level themes.  
However, existing KG methods are primarily designed for simpler document structures, typically limited to \textit{Title} and \textit{Abstract} sections \cite{keyphrase_survey, unikeyphrase}. These methods do not systematically address the complexities of multi-level patent documents, leaving a research gap in their applicability to patent analysis.

To address the challenges of analyzing complex patent documents, we propose KAPPA, \textbf{K}eyphrase-b\textbf{A}sed \textbf{P}ortraits for \textbf{P}atent \textbf{A}nalysis, a novel framework designed to construct interpretable and efficient representations of patents. As depicted in Figure~\ref{fig:kappa}, KAPPA operates in two phases: (1) generating keyphrases and constructing patent portraits (2) leveraging these portraits for downstream analytical tasks, such as patent classification, technology recognition, etc. 

The first phase is the core. For it, we develop an encoder-decoder framework, termed as SetPLM, to generate high-quality keyphrases that capture both salient and latent information within patents. To enhance KG, we adopt the One2Set paradigm \cite{one2set}, which predicts keyphrases in parallel without enforcing a predefined order. This avoids unnecessary biases and improves generation quality. However, One2Set suffers from limitations, including the overestimation of null ($\varnothing$) tokens representing the absence of keyphrases \cite{wrone2set}. To address this, we propose SC-One2Set, which integrates keyword-based calibration to provide semantic guidance during generation.
To exploit the multi-level structure of patents, SetPLM incorporates pre-trained language models (PLMs) for enhanced contextual understanding and prior knowledge archives. While PLMs have been widely utilized in keyphrase extraction \cite{kpeplm} and generation \cite{one2seq, kpgplm}, their integration with One2Set remains unexplored. Our framework demonstrates bidirectional enhancement: PLMs enrich document comprehension for One2Set, while One2Set improves the KG capability of PLMs.
To further leverage the hierarchical nature of patents, we introduce a prompt-based decoding strategy. This involves constructing hierarchical prompts where keyphrases predicted at one level (e.g., \textit{Title}) serve as contextual cues for predicting keyphrases at subsequent levels (e.g., \textit{Abstract} and \textit{Claims}). This approach facilitates efficient information utilization across levels without overwhelming the model with additional textual inputs.
To validate the effectiveness of KAPPA, we conduct extensive experiments on real-world patent datasets. Results demonstrate that our keyphrase-based patent portraits not only provide interpretable and concise representations but also significantly enhance the performance of downstream tasks, including patent classification, technology recognition, and summarization.

This work contributes in the following ways: 
(1) We propose a KG framework to construct patent portraits by extracting both present and absent keyphrases from patent documents. This framework effectively leverages the information within patents while addressing their inherent multi-level structure.  
(2) Building on these portraits, we introduce KAPPA, an analytical framework designed for patent analysis. By utilizing keyphrase-based portraits, KAPPA supports various downstream analytical tasks. These portraits offer a concise yet comprehensive representation of patent documents, enabling efficient processing and improved analytical performance.

The remainder of this paper is organized as follows.  
Section \ref{s:review} reviews the relevant literature.  
Section \ref{s:preliminary} presents the formulation of patent analysis and provides background knowledge on KG. 
Section \ref{s:method} details the proposed methodology, including the framework for KG, portrait construction, and the framework for patent analysis.  
Section \ref{s:experiments} reports extensive experiments evaluating the performance of the proposed method on KG tasks and its effectiveness in enhancing patent analysis tasks.  
Section \ref{s:conclusion} concludes the paper and outlines potential directions for future research.  

\begin{figure}[h]
  \centering
  \includegraphics[width=\linewidth]{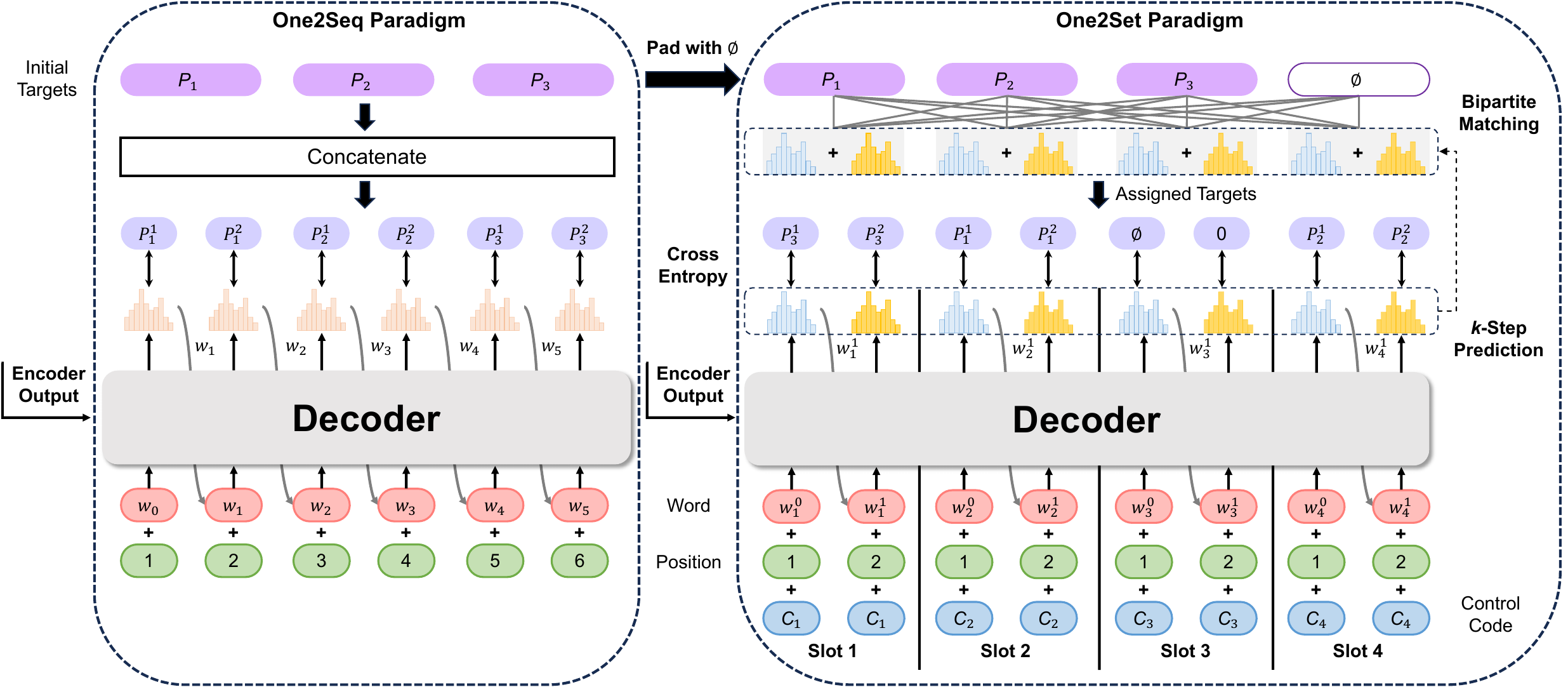}
  \caption{Comparison between One2Seq(left) and One2Set(right). This case focuses mainly on the decoder side and the number of target keyphrases is 3. For One2Set, $N=4$ and $k=2$. }
  \label{fig:one2set}
\end{figure}

\section{Literature Review}
\label{s:review}
In this section, we review the related research and discuss the differences between previous works with ours.
\subsection{Keyphrase Extraction and Generation} 
Methods to obtain keyphrases from documents  can be categorized into extractive and generative paradigms.
Early work focuses on the extractive task \cite{kpe2020,kpeplm}. While extractive methods have salient performance in extracting present keyphrases \cite{kpe2020,kpeplm}, they are incapable of dealing with absent ones which require a comprehensive document understanding \cite{unikeyphrase, promptkp}. Generative methods are designed to tackle this issue by predicting both present and absent keyphrases \cite{kp20k, one2seq, kpgplm}. Generally, these methods fall into three main paradigms: One2One, One2Seq, and One2Set. 
One2One treats each target keyphrase coupled with the document as an independent instance and uses a sequence-to-sequence generative model with a copy mechanism for predicting both present and absent keyphrases \cite{kp20k}. However, One2One can only predict a fixed number of keyphrases. To be more flexible, \cite{one2seq} proposes an One2Seq paradigm, where keyphrases are reordered by a predefined rule, and then concatenated as the input. However, since keyphrases in a document are usually unsorted, imposing a predefined order can potentially introduce unnecessary biases, complicate the training process, and increase sensitivity to the order of keyphrases \cite{one2set, wrone2set}. Addressing these issues, \cite{one2set} proposes a novel One2Set paradigm considering keyphrases as a set instead of a sequence. 
Despite its impressive performance, One2Set still has limitations including overestimation of null ($\varnothing$) tokens that signifies \textit{no corresponding keyphrase}. To solve this problem, \cite{wrone2set} calibrates One2Set with an adaptive instance-level cost weighting strategy and a target re-assignment mechanism. But it demands intricate design and extra elaboration. 

Furthermore, purely generative methods often overlook the semantic relationships between keyphrases and source documents, potentially leading to irrelevant predictions. To constrain generation, \cite{segnet, unikeyphrase} introduce an extractor-generator framework, which leverages present information to guide generation. \cite{promptkp} utilizes keywords to construct prompt templates in a prefix language model. However, these method lacks the capacity for parallel generation due to its reliance on predicting masked tokens, which cannot be expanded to parallel generation. In this work, we also leverage keywords as semantic guides. By adopting the One2Set paradigm, our method enables the parallel generation of keyphrases, making it more efficient than sequential approaches. By directly integrating keywords into the decoder's input, our method achieves more direct and stable constrained generation.

More recently, large language models (LLMs) have been applied to KG tasks. \cite{GPT_keyphrase} and \cite{GPT_survey} demonstrate that ChatGPT can be used for KG, achieving performance comparable to state-of-the-art methods with proper prompt settings. \cite{keyphrase_survey} also validates that ChatGPT outperforms unsupervised keyphrase extraction methods under a zero-shot setting. However, the inference cost of LLMs is high, making it prohibitive to deploy LLMs for KG in large-scale real-world scenarios where the volume of documents is immense or the documents are extremely lengthy, such as in patent analysis. Additionally, \cite{keyphrase_survey} shows that while increasing  parameters of PLMs can be effective, its impact is often overshadowed by the adoption of new training and inference paradigms. Our research addresses this gap by integrating PLMs into the novel One2Set training paradigm. This approach balances enhancing performance through prior knowledge with the implementation costs associated with increasing model parameters.


\subsection{Patent Analysis based on Deep Learning}
Based on deep learning methods for natural language processing, novel approaches have been developed in the field of patent analysis, including patent classification \cite{patent_bert}, patent retrieval \cite{patent_retrieval}, technology forecasting \cite{patent_tkde}, and information extraction and text generation \cite{patent_generation}. For example, \cite{patent_bert} fine-tunes a pre-trained BERT model for patent classification, while \cite{patent_tkde} proposes an event-based framework for patent application trend prediction using graph learning. However, most existing works focus on specific analytical techniques for patent analysis. The development of integrated analysis frameworks based on deep learning methods remains largely unexplored.

Constructing the document representation of patents is the prerequisite of analytical tasks.
\cite{copate} proposes a contrastive learning method to build patent embeddings for analysis. However, the construction of patent embeddings involves semantic compression by simply separating lengthy claims into different parts, thereby neglecting the multi-level features of patents and failing to fully utilize such structural information. Additionally, this document representation is purely numerical and too abstract for experts to explain.
\cite{keyword_patent} extracts keywords from patents to enchance patent classification. \cite{tpe} proposes to extract phrases related to techniques, termed technical phrases, to represent patents. \cite{techpat} constructs a hierarchical framework for patent analysis based on technical phrases. However, these works focus solely on extracting information from words and phrases and do not account for latent information beyond existing texts.
\cite{ai_patent} proposes a novel patent analysis framework based on a language model. However, it primarily focuses on extracting key sentences and phrases from patents, while neglecting extensions to practical downstream patent analysis tasks. Additionally, it requires extra post-training on patents within specific domains, which involves labor-intensive dataset construction and incurs significant costs when transferring to patents across different domains.
Our work is the first to integrate patent structures with both present and absent information in keyphrases, offering a comprehensive representation for patent analysis.

\section{Preliminaries}
\label{s:preliminary}
In this section, we introduce background knowledge of patents and KG. In Section \ref{s:prob_stat}, we provide essential definitions and concepts. In Section \ref{s:one2set}, we discuss how to generate keyphrases as a set and limitations of the existing paradigms.

\subsection{Problem Statement}
\label{s:prob_stat}
\begin{definition}[Multi-level Documents]
A multi-level document is one with multiple levels of texts or sections, i.e., $\mathcal{D} =\left\{\mathcal{X}_1, \mathcal{X}_2, \cdots, \mathcal{X}_{L}\right\}$, where $L$ is the number of levels.
\end{definition}

\begin{definition}[Present and Absent Keyphrases]
From document $\mathcal{D}$, take a document segment $\mathcal{X}$ containing $S$ tokens, i.e., $\mathcal{X}=\left\{x_1, \ldots, x_S\right\}\subseteq\mathcal{D}$, there is a keyphrase set $\mathcal{K}=\left\{K_1, \ldots, K_{|\mathcal{K}|}\right\}$, where each keyphrase $K_i (i=1,\cdots, |\mathcal{K}|)$ is a sequence of tokens.
$\mathcal{K}$ can be 
 further divided into a present keyphrase set $\mathcal{K^P}$ and an absent keyphrase set $\mathcal{K^A}$, where $\mathcal{K^P}\cap \mathcal{K^A} = \varnothing$. In particular, 
 $\forall K^P \in \mathcal{K^P}$, $K^P\cap\mathcal{X}=K^P$; 
 $\forall K^A \in \mathcal{K^A}$,   $K^A\cap\mathcal{X}\subset K^A$. 
\end{definition}

\begin{definition}[Keywords]
\label{def:keyword}
Given a document segment $\mathcal{X}$, keywords are sequences of tokens that shared by keyphrases $\mathcal{K}$ and $\mathcal{X}$, noted as $\mathcal{W^K} = \mathcal{K}\cap \mathcal{X}$, where $\mathcal{W^K}$ is the keyword set of $\mathcal{X}$. 
\end{definition}

Given a patent including \textit{Title}, \textit{Abstract}, and \textit{Claims} sections, we formulate it as $\mathcal{D} =\left\{\mathcal{X}_1, \mathcal{X}_2, \mathcal{X}_3\right\}$. In practice, we combine several short sections (e.g., \textit{Title} and \textit{Abstract}) for comprehensive information; while we have to split a long text into several parts for machine to understand (e.g., \textit{Claims}). Therefore, we rewrite a patent as $\mathcal{D} =\left\{\mathcal{X}_1, \mathcal{X}_2, \cdots, \mathcal{X}_{C+1}\right\}$, where $\mathcal{X}_1$ is the concatenation of Title and Abstract, and $C$ is the number of text segments for Claims after separation. For simplicity, we term each segment directly fed into the model as a document segment, noted as $\mathcal{X}$. We note the keyphrase set and keyword set of the entire patent as $\mathcal{K} = \bigcup\limits_{l=1}^{C+1}\mathcal{K}_l$ and $\mathcal{W}^{\mathcal K} = \bigcup\limits_{l=1}^{C+1}\mathcal{W}^{\mathcal{K}}_l$ respectively, where $\mathcal{K}_l$ and $\mathcal{W}^\mathcal{K}_l$ are the keyphrase set and the keyword set at level $l$. In this paper, we use $\mathcal{K}$ as the portrait of patent $\mathcal{D}$. 





\subsection{One2Set: Generate Keyphrases as a Set} 
\label{s:one2set}
\noindent One2Set is originally implemented with a Transformer-based model \cite{one2set}. As shown in Figure\ref{fig:one2set}, One2Set first appends target keyphrases to a list $\mathcal{T}$ of size $N$. If the number of targets is smaller than $N$, the vacancies are filled with null($\varnothing$) tokens which indicate \textit{no corresponding keyphrases}.
To realize parallel generation, One2Set divides  decoder inputs into $N$ slots and introduces $N$ control codes as additional decoder inputs for slot distinction. During training, each slot learns to generate a keyphrase or a $\varnothing$ token. 

\textbf{\textit{$K$-Step Target Assignment Mechanism.}} 
In One2Set, correspondence between each prediction and its target is unknown, since there is no predefined order of keyphrases. 
To address this issue, SetTRANS employs a $k$-step target assignment mechanism to arrange targets to predictions. Specifically, decoder first conducts a $k$-step prediction and generates $k \times N$ probability distributions to describe the first $k$ tokens predicted by $N$ slots.
Based on the first $k$ tokens of each target and prediction, a bipartite matching algorithm \cite{matching} searches for a permutation policy of targets. 
The optimal policy $\pi^*$ is defined as follows:
\begin{equation}
\small
\pi^*=\mathop{\rm{argmin}}\limits_{\pi\in\Pi(N)}\mathop\sum_{n=1}^N\mathop\sum_{t=1}^{f(k)}\mathbb{I}({\mathcal T}^t_{\pi(n)}\neq\varnothing) p^t_n({\mathcal T}^t_{\pi(n)})
\label{eq:kstep} 
\end{equation}
where $f(k)=\min(k, \lvert {\mathcal T}_{\pi(n)}\rvert)$. $\Pi(N)$ refers to all possible schemes of permuting the $N$-length target list $\mathcal{T}$; while $\pi(n)$ is the index of the target that $\pi$ assigns to slot $n$. 
Further, $p^t_n({\mathcal T}^t_{\pi(n)})$ is the probability of token ${\mathcal T}_{\pi(n),t}$ in  predicted distribution $p^t_n$; while $\mathbb{I}(\cdot)$ is an indicative function that merely takes non-$\varnothing$ targets into consideration. Finally, each target will be assigned an one-and-only prediction, and the model is trained with a loss function based on cross-entropy. 

\textbf{\textit{Mis-calibration: $\varnothing$-Token Overestimation($\varnothing$-OV).}}
One major limitation of One2Set is that some slots which are able to generate keyphrases actually generate $\varnothing$ tokens, termed $\varnothing$-OV. As pointed out by \cite{wrone2set}, during inference, with $\varnothing$ tokens removed, $14.6\%$ of slots predict correct keyphrases; while $34.5\%$ of them directly predict $\varnothing$ tokens without removal. This issue mainly stems from One2Set's padding strategy, where $\varnothing$ tokens are used to fill the target list $\mathcal{T}$ to maintain a fixed size of $N$. 
First, the excessive $\varnothing$ tokens later serve as supervised signals during training.
Second, the prevalence of $\varnothing$ tokens introduces instability in the $k$-step target assignment: since they are not considered in computing the matching policy in Equation \eqref{eq:kstep}, the same slot may be assigned either a keyphrase or the $\varnothing$ token in different iterations. This further aggravates $\varnothing$-OV. 

\section{Methodology}
\label{s:method}
In this section, we discuss our method of KG and the framework utilizing keyphrases for patent analysis. In Section \ref{s:sc_one2set}, we introduce SC-One2Set, a semantic-calibrated generation scheme based on keywords. In Section \ref{s:setplm}, we propose SetPLM, a paradigm that incorporates PLMs to enhance KG using the prior knowledge contained in PLMs. In Section \ref{s:kappa}, we present KAPPA, a patent analysis framework based on portraits constructed from keyphrases.

\begin{figure}[!ht]
  \centering
  \includegraphics[scale=0.4]{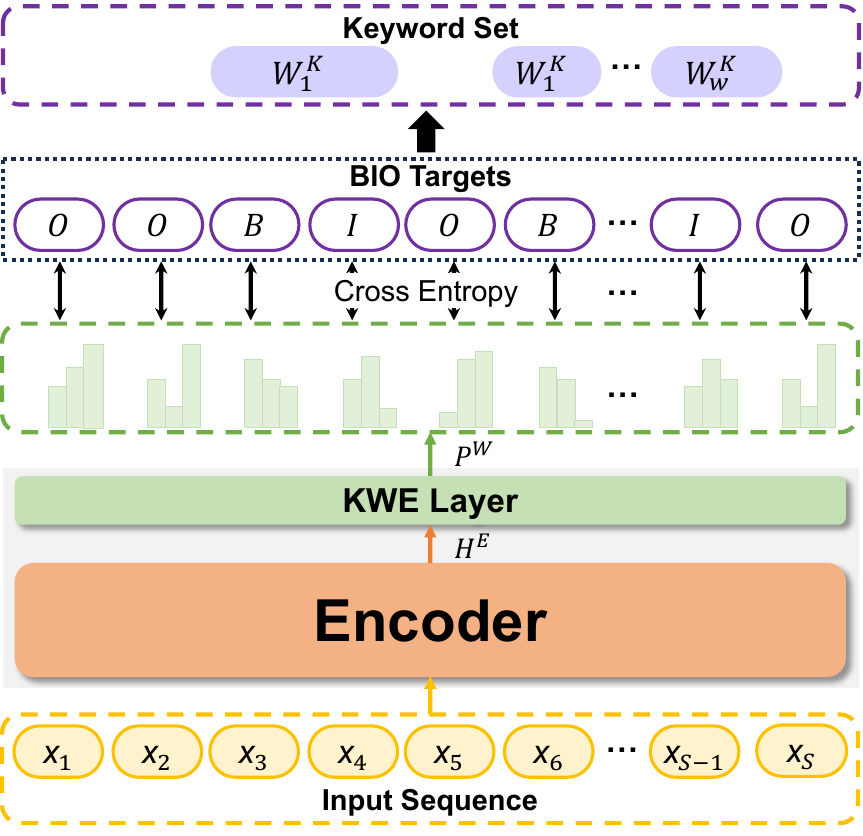}
  \caption{Keyword Extraction on the Encoder Side of SC-One2Set.}
  \label{fig: encoder}
\end{figure}

\subsection{SC-One2Set: Semantic-Calibrated Keyphrase Generation}
\label{s:sc_one2set}
To obtain high-quality keyphrases, we aim to take advantages of One2Set paradigm and mitigate its limitations. 

The first is to alleviate $\varnothing$-OV. Since $\varnothing$-OV is mainly caused by the padding strategy with $\varnothing$ tokens, one intuition to alleviate that is to reduce the portion of $\varnothing$ tokens in the target list $\mathcal{T}$. 
To achieve this, WR-One2Set\cite{wrone2set} employs a target re-assignment mechanism. It records slots assigned $\varnothing$ tokens during $k$-step target assignment, removes $\varnothing$ from the vocabulary, and performs $k$-step prediction again. Among recorded slots, those with unique predictions receive the best-matched non-$\varnothing$ targets from $\mathcal{T}$; while those with predictions identical to others receive no target. Despite its effectiveness, there are some issues required to be solved. 
First, the re-assignment mechanism will assign the same targets to different slots, thus significantly increase duplicated predictions. Second, during each training iteration, it performs the $k$-step prediction and assignment twice. This process greatly impacts efficiency of the algorithm.
Third, performing non-$\varnothing$ prediction is unstable because removing the $\varnothing$ token damages the integrity of vocabulary, thus introducing bias to model training.

The second is to make the conditional generation more controllable. As depicted in Figure \ref{fig:one2set}, the initial token of each slot $w^1_n$ holds significance. During the training stage, $k$-step target assignment relies on the first $k$ tokens predicted by each slot, with the prediction heavily influenced by its initial token. Consequently, $w^1_n$ plays a crucial role in determining the training target for slot $n$. Similarly, during the inference stage, predictions for slot $n$ also heavily depend on $w^1_n$. However, in vanilla One2Set, $w^1_n$ is determined by the control code for slot $n$, $C_n$, along with the position embedding and the start token for time step $1$, $P^1$ and $w^0_n$, respectively. Notably, $P^1$ and $w^0_n$ are identical for each slot. This renders the control mechanism uncontrollable. To be specific, for the same document, different slots only differ in their control codes when generating their first tokens; while for different documents, the same control code for the same slot is applied across distinct inputs.

These limitations not only compromise the generation quality of One2Set, but also hinder enhancement of extensive knowledge from PLMs, as they become more pronounced with the integration of PLMs. Hence, we introduce keywords as a promising semantic representation for calibration. Our method, SC-One2Set, comprises two steps: keyword extraction (KWE) and keyword-based KG. Leveraging the encoder-decoder transformer-based model structure from \cite{one2set, wrone2set}, we implement KWE on the encoder side and KG on the decoder side to further maximize its potential.

\subsubsection{Keyword Extraction (KWE)} 
Referring to Definition \ref{def:keyword}, we begin by extracting overlapping words between documents and their ground-truth keyphrases to construct our ground-truth keyword set $\mathcal{\hat{W}^K}$. This approach is necessary as benchmark datasets for KG typically provide ground-truth keyphrases exclusively.

Next, we frame KWE as a sequence labeling task, aiming to predict labels $y^W = \left[y^W_1, y^W_2, \cdots, y^W_S\right]$ for the input sequence $I^E = \left[x_1, x_2, \cdots, x_S\right]$. Following the convention of \cite{unikeyphrase}, we adopt a BIO format classification, where $y^W_s \in \mathcal{C}$ and $\mathcal{C}=\{B, I, O\}$ for $s=1,2, \cdots, S$. In this format, $B$ denotes the first token of a keyword, $I$ signifies the other components of a keyword, and tokens without a keyword label are marked with $O$. Utilizing this scheme, we construct the target sequence by labeling each token of the input sequence in the BIO format.

As shown in Figure \ref{fig: encoder}, we feed $I^E$ into the encoder parameterized with $\theta^E$ and get the last hidden state $H^E$ as follows: 
\begin{align*}
    H^E &= \theta^E(I^E) = \left[h^E_1, h^E_2, \cdots, h^E_S\right], \quad h^E_s \in \mathbb{R}^{d\times 1}, 
\end{align*}
where $d$ is the output dimension of the encoder.

Subsequently, a classifier is added on top of the encoder. This classifier, termed the KWE layer, comprises a fully connected layer with an input dimension of $d$ and an output dimension of $|\mathcal{C}|$, along with a SoftMax activation function. We feed $H^E$ into the KWE layer to predict the probability of each token $x_s$ in $I^E$ being a component of a keyword in the BIO format, denoted as:
\begin{align*}
    p^W &= \text{KWE}(H^E) = \left[p^W_1, p^W_2, \cdots, p^W_S\right], \quad p^W_s\in \mathbb{R}^{|\mathcal{C}|\times1}
\end{align*}

In particular, we can formulate the relationship between $h^E_s$ and $p^W_s$ as follows: 
\begin{equation}
    p^W_s = {\rm softmax}\left(\left(\omega^W h^E_s + \beta^K\right)\right), 
\end{equation}
where $\omega^W$ and $\beta^W$ are learnable parameters of the KWE layer.
For dimension $i$ of $p^W_s$, we write $p^W_{s,i}$ as follows: 
\begin{align}
    p^W_{s,i} &= P_r(c_i\mid H^E), \quad c_i \in \mathcal{C}
\end{align} 

Therefore, the predicted category of token $x_s$ is:
\begin{equation}
    y^W_s = \mathop{\rm argmax}\limits_{c_i \in \mathcal{C}}{p^W_{s,i}}
\end{equation}
Based on KWE, we collect tokens of which $y^W_s\in\{B, I\}$ and construct the keyword set $\mathcal{W^K}$. 

\begin{figure}[!t]
  \centering
  \includegraphics[width=\linewidth]{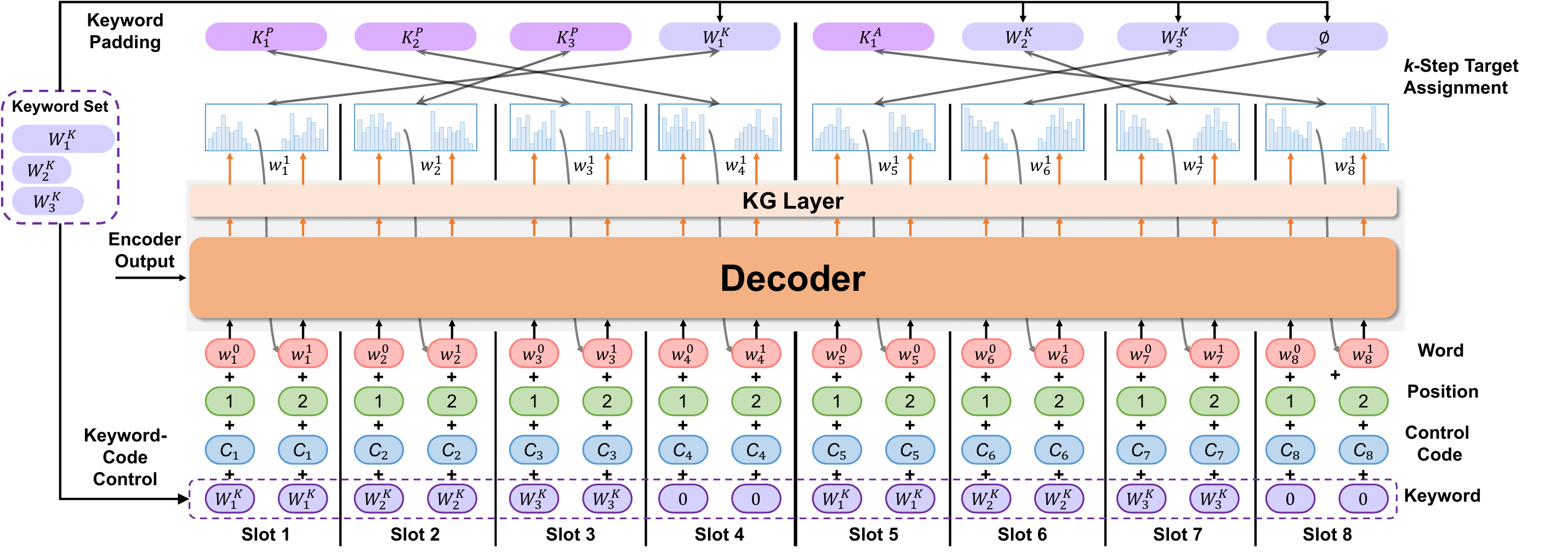}
  \caption{The framework of SC-One2Set. In this case, $N=8$ and $k=2$. For simplicity, we use a special case where $\mathcal{W^K_P} = \mathcal{W^K}$. }
  \label{fig: setplm}
\end{figure}

\subsubsection{Keyword-based Keyphrase Generation}
Following \cite{one2set}, we consider the target keyphrases as a set of size $N$ and separate the decoder inputs into $N$ slots, where $N/2$ for present keyphrases and where $N/2$ for absent ones. In particular, we propose Keyword Padding (KWP) Strategy and Keyword-Code Control (KCC) Mechanism to incorporate keywords into KG. 

\textbf{Keyword Padding (KWP) Strategy.} 
To prevent an excess of $\varnothing$ tokens in targets, the KWP strategy pads targets with keywords from $\mathcal{W^K}$. As some present keyphrases are single words (collected as $\mathcal{\hat K}^P_S$), we eliminate keywords shared by $\mathcal{\hat K}^P_S$ and $\mathcal{W^K}$ from $\mathcal{W^K}$ to avoid introducing duplicated targets and build a padding keyword set $\mathcal{W^K_P}$, denoted as: 
\begin{equation*}
    \mathcal{W^K_P}\leftarrow \mathcal{W^K}-\mathcal{W^K}\cap\mathcal{\hat K}^P_S
\end{equation*}
In Figure \ref{fig: setplm}, $\mathcal{W^K_P}=\{W^K_1, W^K_2, W^K_3\}$ provides keywords for target padding, ensuring no overlap with existing ground-truth keyphrases. Subsequently, we append the ground-truth keyphrase set $\mathcal{\hat K}$ and $\mathcal{W^K_P}$ to the target list $\mathcal{T}$. In scenarios where the combined count of keyphrases and keywords is insufficient to meet the desired length of $\mathcal{T}$ (i.e., $|\mathcal{\hat K}|+|\mathcal{W^K}| < N$), vacancies are filled with $\varnothing$ tokens. 

By replacing a significant portion of the $\varnothing$ tokens with keywords, KWP introduces increased diversity of targets, thus reduced $\varnothing$-OV.
During the training stage, the KWP strategy enables  a direct utilization of the conventional $k$-step target assignment in Equation \eqref{eq:kstep} and allows each slot to be assigned a high-quality target sampled directly from $\mathcal{T}$ without the need for redundant non-$\varnothing$ predictions and the laborious target re-assignment employed by \cite{wrone2set}. Since these keywords differ from the existing ground-truth keyphrases, their inclusion prevents additional duplicated predictions.
In the inference stage, we eliminate predictions identical to those in $\mathcal{W^K_P}$ and construct the final keyphrase set $\mathcal{K}$.

\textbf{Keyword-Code Control (KCC) Mechanism.} 
To integrate semantic guidance, we design a KCC mechanism to control generation. 
As shown in Figure \ref{fig: setplm}, KCC first constructs embedding of keywords extracted from $\mathcal{W^K}$. For each slot, KCC adds embeddings of keyword and control codes of to construct a new control embedding $e(C^W_n)$, denoted as:
\begin{equation}
    e(C^W_n) = e({C}_n) + e({W}^K_{\delta(n)}),\quad {W}^K_{\delta(n)}\in \mathcal{W^K}
\end{equation}
where $e(C_n)$ and $e(W^K_{\delta(n)})$ are embeddings for the control code and the keyword of slot $n$, respectively. Here, $\delta(n)\in\{1,2,\cdots,N^K\}$ is an index of the keyword in $\mathcal{W^K}$. Particularly, we use $N^K$ keywords from $\mathcal{W^K}$ and $N^K<{N/2}$.  
First, by combining with different codes, the same $\mathcal{W^K}$ forms distinct control embeddings, effectively distinguishing slots. Given that KWP introduces padding keywords, which may partially overlap with the original ground-truth keyphrases. This also maintains semantic relevance between slots that share the same $\mathcal{W^K}$, ensuring effective collaboration with KWP. Moreover, KCC is designed to guide generation with semantic information rather than imposing a rigid guarantee. Excessive keywords may introduce noise, as not every keyphrase is relevant to a keyword, especially the absent ones. Therefore, we opt for the best $N^K$ keywords predicted by KWE instead of filling all slots with unique keywords.

The decoder's input for slot $n$ at time step $t$ is as follows: 
\begin{equation}
    {I^D_n}^t = e(w^{t-1}_n) + e(P^t) + e(C^W_n) 
\label{eq:decoder_inputs}
\end{equation}
where $e(w^{t-1}_n)$ denotes the token embedding of slot $n$ generated at time step $t-1$ and $w^0_n$ is the start token for decoding, e.g., \textit{[BOS]} or \textit{[PAD]}. 
$e(P^t)$ represents the position embedding for the token generated at time step $t$ and $P^t=t$. Notably, $t=1,2,\cdots,T$, with $T=k$ during $k$-step target assignment and $T=|\mathcal{T}_{\pi^*(n)}|$ during training. 

The decoder's last hidden state $H^D$ is as follows: 
\begin{equation*}
    H^D=[h^D_1, h^D_2, \cdots, h^D_N], \quad h^D_n\in \mathbb{R}^{d\times m}
\end{equation*}
Subsequently, the decoder feeds $H^D$ into the KG layer, comprising a fully connected layer with an input dimension of $d$ and an output dimension of $|\mathcal{V}|$, where $\mathcal{V}$ denotes the vocabulary. This layer is accompanied by a SoftMax activation function, and its output represents the predicted probability distribution for each token in the vocabulary $\mathcal{V}$. 

At time step $t$, output of the KG layer is as follows:
\begin{equation}
    p^t_n = {\rm softmax}(\omega^G {h^D_n}^t + \beta^G)
\end{equation}
where ${h^D_n}^t$ is the decoder's last hidden state for slot $n$ at time step $t$, while $\omega^G$ and $\beta^G$ are learnable parameters for the KG layer. 
Specifically, prediction for this step, $w^t_n$, is the token with the highest probability, expressed as follows: 
\begin{equation}
    w^t_n = {\rm \mathop{argmax}\limits_{\textit{w}\in \mathcal{V}}}\left(p^t_n(w)\right)
\end{equation}
    
With KCC, we incorporate semantic information into the decoder's inputs, thus providing enhanced control by distinguishing slots during the parallel generation and improving generalization across different documents through prior information related to the input. 

\subsection{SetPLM: Incorporating a PLM to SC-One2Set }
\label{s:setplm}
\noindent
With semantic calibration, SC-One2Set overcomes limitations of vanilla One2Set, thus becoming scalable to PLMs. We term this scheme as SetPLM. 

\subsubsection{Model Structuring}
Without loss of generality, we select a commonly-used PLM, T5 \cite{t5}, as the backbone PLM of SetPLM. T5 is known for its exceptional performance across diverse text-based tasks, notably including KG \cite{t5, plms4kg}. Furthermore, given its encoder-decoder structure, T5 is conveniently transplanted to SC-One2Set. 

\textbf{Dual-Ordering Position Embedding (DOPE).} 
In SC-One2Set, two types of orders exist: keyphrases are unordered, while tokens within each keyphrase follow a specific order, which is periodic by slot.
As PLMs are commonly pretrained with a certain type of position embedding \cite{t5}, we introduce the DOPE to guide the decoder distinguish between these two orders. 
In SetPLM, DOPE first incorporates an absolute positional embedding (APE), assigning a $d$-dimensional positional vector $e(P^t)$ to token $t$ of each slot and adding it to $I^D$. 
Subsequently, as in vanilla T5, DOPE employs a bucketing-based relative positional embedding (RPE) in its attention layer to model the positional difference between tokens $u$ and $v$ within $I^D$, denoted as $\rho(u ,v)$. Hence, the attention weight $\text{AW}(u, v)$ is as follows:
\begin{equation}
\text{AW}(u, v)=\frac{\left(I^D(u) \Omega^Q\right)\left(I^D(v) \Omega^K\right)^T+\rho(u ,v)}{\sqrt{d}}, 
\end{equation}
where $\Omega^Q $ and $\Omega^K$ are parameters of the query and key matrices. Since RPE is conducted in the attention layer, ${I^D}$ maintains the format in Equation \eqref{eq:decoder_inputs}, where $e(P^t)$ with APE can be further written by dimension in a sinusoidal format as follows:
\begin{align}
e(P^t)_{2i} & =\sin \left(\frac{P^t}{10000^{2 i / d}}\right) \notag\\ 
e(P^t)_{2i+1} & =\cos \left(\frac{P^t}{10000^{2 i / d}}\right),\quad i \in [1,\dfrac{d}{2}]
\end{align}

\textbf{Semantic Compression for KCC.} 
Vanilla One2Set utilizes a WordPiece tokenizer and constructs the vocabulary from the training dataset. In contrast, for enhanced generalization, SetPLM adopts the vocabulary of the PLM and the accompanying pretrained SentencePiece tokenizer. Consequently, words may be tokenized into several sub-words. To stabilize KCC, we compress the embedding of keywords, ensuring complete semantic guidance for each slot. This process is denoted as follows:
\begin{equation}
    e(W^K_n) = \mathop{\sum}\limits_{q}^{Q_n} {e(\boldsymbol{W^K_n})_q}
\end{equation}
Here, $Q_n$ represents the number of tokens contained in the initially tokenized $\boldsymbol{W^K_n}$.

\subsubsection{Three-Stage Multi-Task (TSMT) Training} 
In our proposed SC-One2Set, we need to train the model on KWE and KG tasks. However, training tasks separately is inefficient and overlooks their interdependence, while simultaneous training of encoder and decoder on different tasks is unstable, prone to overfitting—especially when incorporating PLMs with a large number of parameters. To deal with these problems, we design a novel TSMT training scheme for SC-One2Set and SetPLM. 

\textbf{Stage 1: Training Encoder on KWE.} 
At this stage, we fix parameters of the decoder and train the encoder independently on KWE for $\mathcal{E}_1$ epochs.
We denote the loss function $\mathcal{L}^1_{W}$ as follows:
\begin{align}
    \mathcal{L}^1_{W} &= -\frac{1}{S}\mathop\sum_{s=1}^{S}\mathop\sum_{c_i \in \mathcal{C}}\xi^W_{c_i} \hat {y}_{s,i}^{W} \log\left({p}_{s,i}^{W}\right),
\label{eq:lw1}
\end{align}
where $\xi^W_{c_i}$ is the weights of category $c_i$ when calculating loss. Since most words of a document are non-key words, which will be classified to \textit{O} in KWE, we set the weights with the reciprocals of numbers of the corresponding categories to alleviate imbalance of classification. $\hat {y}_{s,i}^{W}$ refer to dimension $i$ of the spanned ground truth for prediction ${p}_s^W$.  

\textbf{Stage 2: Training Decoder on KG.} 
At this stage, we fix parameters of the encoder and train the deocder on KG for $\mathcal{E}_2$ inner epochs. 
Based on $k$-step target assignment described by Equation \eqref{eq:kstep}, the loss function for KG task is formulated as follows:
\begin{equation}  
    \mathcal{L}_{G}=-\mathop\sum_{n=1}^{N}\mathop\sum_{t=1}^{\lvert \mathcal{T}_{{\pi^*}(n)}\rvert}{\xi^G}(\pi^*(n))\log {\overline p}^t_{n}\left(\mathcal{T}^t_{\pi^*(n)}\right)
\label{eq:lg}
\end{equation}
where ${\overline p}^t_n$ is the $t$-th predicted probability distribution of the $n$-th slot using teacher forcing \cite{one2set, wrone2set}. Here, ${\xi^G}(n)$ is an adaptive weight scalar used to reduce the class imbalance of keyphrases, keywords and $\varnothing$ tokens, noted as follows:
\begin{equation*}
{\xi^G}(\pi^*(n))
= \begin{cases}
\lambda_{\varnothing},&{\text{if}}\ T_{\pi^*(n)} = \varnothing \\ 
\lambda_{W},&{\text{if}}\ T_{\pi^*(n)} \in \mathcal{W^K} \\
0,& \text{otherwise}
\end{cases}
\end{equation*}
$\lambda_{\varnothing}$ and $\lambda_{W}$ are hyper-parameters.

\textbf{Stage 3: Tuning Encoder for KG.} 
Since we aim to generate high-quality keyphrases at last, the model should understand the document not only based on KWE, but also the KG. Therefore, we fix the decoder's parameters and further tune the encoder's parameters based on the performance of KG at this stage. 
The loss function $\mathcal{L}^2_W$ is as follows:
\begin{equation}
    \mathcal{L}^2_W = \mathcal{L}^1_{W} + \lambda_G\dfrac{1}{{\mathcal{E}}_2}\sum\limits_{\epsilon=1}^{\mathcal{E}_2}\mathcal{L}_{G}^{\epsilon}
\label{eq:lw2}
\end{equation}
where $\lambda_G$ is the weight of KG performance for the encoder, $\mathcal{L}_{G}^{\epsilon}$ is the loss of the $\epsilon$-th inner epoch during stage 2. 

Algorithm \ref{alg:tmst} outlines the overall framework of TSMT. Here, for simplicity, we use a single document segment $\mathcal{X}$, so as to emphasize the high-level training scheme.

\begin{algorithm}[!t]
\begin{algorithmic}[1]
\caption{TSMT for SC-One2Set}
\label{alg:tmst}
\REQUIRE Document segment $\mathcal{X}$, ground-truth keywords and keyphrases
\ENSURE Keyphrase set $\mathcal{K}$ from $\mathcal{X}$
\STATE \textbf{Initialize:} 
PLM's parameters $\{\theta^E, \theta^D\}$, KWE and KG layers' parameters $\{\theta^W, \theta^G \}$
\FOR{epoch $\in \{1,2, \cdots, \mathcal{E}\}$}
\IF{epoch $\leq \mathcal{E}_1$}
\STATE Compute $\mathcal{L}^1_W$ with Equation \eqref{eq:lw1}
\STATE Update $\theta^E$ with: $\theta^E \leftarrow \theta^E-\alpha_W \nabla_{\theta^E} \mathcal{L}^1_{W}$
\STATE Update $\theta^W$ with: $\theta^W \leftarrow \theta^W-\alpha_W \nabla_{\theta^W} \mathcal{L}^1_{W}$
\ELSE
\STATE $\mathcal{S}_{G} \leftarrow 0$
\STATE Conduct KWE and construct the keyword set $\mathcal{W^K}$. 
\STATE Build target list $\mathcal{T}$ and conduct KWP with $\mathcal{W^K}$. 
\FOR{inner epoch $\in \{1,2, \cdots, \mathcal{E}_2\}$}
\STATE Assign targets to each slot, including $k$-step prediction and target assignment based on Equation\eqref{eq:kstep}.
\STATE Compute $\mathcal{L}_G$ with Equation \eqref{eq:lg}
\STATE Update $\theta^D$ with: $\theta^D \leftarrow \theta^D-\alpha_G \nabla_{\theta^D} {\mathcal{L}}_{G}$
\STATE Update $\theta^G$ with: $\theta^G \leftarrow \theta^G-\alpha_G \nabla_{\theta^G} {\mathcal{L}}_{G}$
\STATE $\mathcal{S}_{G} \leftarrow \mathcal{S}_{G}  + {\mathcal{L}}_{G}$
\ENDFOR 
\STATE Compute $\mathcal{L}^2_W$ with Equation \eqref{eq:lw2}
\STATE Update $\theta^E$ with: $\theta^E \leftarrow \theta^E-\alpha_W \nabla_{\theta^E} {\mathcal{L}^2_W}$
\STATE Update $\theta^W$ with: $\theta^W \leftarrow \theta^W-\alpha_W \nabla_{\theta^W} {\mathcal{L}^2_W}$
\ENDIF
\ENDFOR
\end{algorithmic}
\end{algorithm}

\subsection{KAPPA: Keyphrase-bAsed Portraits for Patent Analysis}
\label{s:kappa}
KAPPA comprises two phases: portrait generation and portrait-based patent analysis. The first phase involves generating keyphrases that account for the structural features of patents and constructing patent portraits based on these keyphrases, incorporating domain knowledge. The second phase involves conducting practical analytical tasks related to intellectual property using the generated patent portraits.
\subsubsection{Phase 1: Generating Keyphrase-based Portraits}
To construct patents' portraits based on keyphrases, here are two challenges. First, it is non-trivial to predict keyphrases of patents in a single step due to their extensive lengths. Dividing patents into segments and predicting keyphrases separately may result in a loss of relevance, consequently limiting the quality of keyphrases. Second, patents are scientific documents relevant to various domains, and it is necessary to consider domain-specific knowledge when constructing portraits. To address these two issues, we introduce a prompt-based hierarchical decoding (PHD) strategy.

Prompting is a technique that involves adding extra information for the model to condition on during generation, and has been widely used in previous works to enhance the inputs of language models \cite{t5, promptkp}. Specifically, we denote prompting as $P_r(Y|[\mathcal{X}^p;\mathcal{X}])$, where $\mathcal{X}^p$ represents a series of tokens in the prompt prepended to $\mathcal{X}$. The model then maximizes the likelihood with fixed parameters. For example, for a single document segment $\mathcal{X}$ of normal length, we can directly utilize a task-related prefix to construct $\mathcal{X}^p$. In the case of KG, the prefix is written as \textit{find keyphrases from:}.

In KAPPA, we aim to utilize prompting to effectively handle complex inputs, such as patents. For a multi-level document $\mathcal{D} = \{\mathcal{X}_1, \mathcal{X}_2, \cdots, \mathcal{X}_{C+1}\}$, PHD constructs $\mathcal{X}^p_l$ for each document segment $\mathcal{X}_l\in \mathcal{D}$ using keyphrases from the previous levels. Specifically, the input sequence $[\mathcal{X}^p_l; \mathcal{X}_l]$ is defined with the template: \textit{keyphrases from higher-level: \{\(\mathcal{K}_{l-1}\)\} [sep] find keyphrases from: \{\(\mathcal{X}_l\)\}}. For the first level, i.e., $l=1$, we use the keyword set for this level as $\mathcal{K}_0$, denoted as $\mathcal{K}_0 = \mathcal{W}^\mathcal{K}_1$. 
For each patent, we obtain a set of keyphrases, $\mathcal{K}$, generated from different segments of $\mathcal{D}$. We term $\mathcal{K}$ as the portrait of patent $\mathcal{D}$. Compared to keyphrases generated directly from $\mathcal{D}$, the portrait $\mathcal{K}$ offers two advantages. First, the construction of $\mathcal{K}$ leverages information from each level of the document. For example, the \textit{Title} typically contains information strongly related to the domain and expresses it concisely. Therefore, $\mathcal{K}_0$ always includes domain-specific information and serves as a foundation for prompting KG in subsequent levels. During this process, $\mathcal{K}$ incorporates domain-specific knowledge. This is essential for analyzing documents from various domains that require prior knowledge. Second, unlike previous works that consider the multi-level structure of patents, PHD does not rely on pre-existing domain-specific information, such as IPC/CPC codes used in \cite{techpat}, making it promising for analytical cases where prior information is not available.
Algorithm \ref{alg:phd} outlines the procedure of PHD. 

\begin{algorithm}[!t]
\begin{algorithmic}[1]
\caption{Prompt-based Hierarchical Decoding Strategy}
\label{alg:phd}
\REQUIRE A document $\mathcal{D} =\left\{\mathcal{X}_1, \mathcal{X}_2, \cdots, \mathcal{X}_{C+1}\right\}$, SetPLM's tuned parameters $\hat\theta = \{\hat\theta^E, \hat\theta^D\}$ with TSMT
\ENSURE A keyphrase set $\mathcal{K}$ of document $\mathcal{D}$ 
\STATE \textbf{Initialize:} 
$\mathcal{K} = \varnothing$. 
\FOR{level $l \in \{1,2, \cdots, C+1\}$}
\STATE KWE with $\hat\theta^E$ and construct $\mathcal{W}^\mathcal{K}_l$
\IF{$l=1$}
\STATE $\mathcal{K}_{l-1}=\mathcal{W}^\mathcal{K}_l$
\ENDIF
\STATE Construct prompt $\mathcal{X}^p_l$ with $\mathcal{K}_{l-1}$
\STATE Construct the input sequence $\mathcal{X}_l=[\mathcal{X}^p_l;\mathcal{X}_l]$
\STATE KG with $\hat\theta^D$ based on $\mathcal{W}^\mathcal{K}_l$ and construct $\mathcal{K}_l$
\STATE $\mathcal{K}\leftarrow \mathcal{K}\cup\mathcal{K}_l$
\ENDFOR
\end{algorithmic}
\end{algorithm}

\subsubsection{Phase 2: Patent Analysis based on Portraits}
In this phase, we apply portraits for patent analysis. We introduce two modes of incorporating portraits to patent analysis. The first is \textit{Pure Portraits}, which involves using only keyphrases. This mode assesses the representative ability of the portraits. The second mode is \textit{Portrait Augmentation}, which involves adding portraits to the original texts as an augmentation. This mode evaluates the semantic supplementary information provided by the portraits and how it can be used to enhance the original texts.
Without loss of generality, we choose three typical downstream patent analysis tasks: patent classification, technology recognition, and patent summerization.

In the field of intellectual property, patent classification aims to predict the IPC/CPC codes of a patent application given a subset of the patent's text. This automatic classification of patents into various technical fields facilitates the effective assignment of patent applications to examiners, improving the efficiency and accuracy of the examination process \cite{suzgun2024harvard}.
Technology recognition involves recognizing the technologies contained within patents. Technologies are defined by phrases that are closely related to the core technology of the patents. Unlike the technical phrases defined by \cite{techpat, tpe}, which can be directly extracted from the patent texts, technologies may not be explicitly present in the text. Therefore, technology recognition requires a more abstract understanding of the latent representations of a patent. Compared to CPC codes, technologies are more high-level and can be defined to meet demands related to specific topics that patent experts may be involved in.
Patent summarization involves creating an abstract based on a subset of the text to generate a summarization of the patent. Following \cite{suzgun2024harvard}, we input the \textit{Claims} section to a baseline generative. To validate the semantic value added by portraits, we also combine portraits with the original texts as \textit{Portrait Augmentation}. 

\section{Experiments}
\label{s:experiments}
In this section, we conduct extensive experiments to evaluate the KG performance of SetPLM and the effectiveness of applying KAPPA framework to patent analysis.
In Section \ref{s:exp_settings}, we discuss the experimental settings. In Section \ref{s:result}, we analyze the results of our experiments. 

\begin{table*}[!t]
\caption{KG Performance on Benchmark Datasets.}
\centering
\begin{tabular*}{\linewidth}{@{\extracolsep{\fill}}l|l|cc|cc|cc|cc|cr}
\toprule
\multirow{2}{*}{\textbf{Task\tnote{1}}}&
\multirow{2}{*}{\textbf{Model}}&
\multicolumn{2}{c|}{\textbf{KP20K}}& \multicolumn{2}{c|}{\textbf{Inspec}}& \multicolumn{2}{c|}{\textbf{Krapivin}}& \multicolumn{2}{c|}{\textbf{NUS}}&
\multicolumn{2}{c}{\textbf{SemEval}}\\
&&
\textbf{$F_1@5$} & \textbf{\textbf{$F_1@M$}}&
\textbf{$F_1@5$} & \textbf{\textbf{$F_1@M$}}& 
\textbf{$F_1@5$} & \textbf{\textbf{$F_1@M$}}&
\textbf{$F_1@5$} & \textbf{\textbf{$F_1@M$}}&
\textbf{$F_1@5$} & \textbf{\textbf{$F_1@M$}}\\
\midrule
\multirow{6}{*}{\makecell[cl]{Present\\Keyphrases}}
&UniKeyphrase
&0.347&0.352
&0.260 &0.288
&-&-
& 0.415&0.443
&0.302&0.322\\
&PromptKG 
&0.351&0.355
&0.260&0.294
&-&-
& 0.412&0.439
&0.329&0.356\\
&SetTRANS 
& 0.358&\underline{0.392}
&0.285&0.324
&0.326&0.364
& 0.406&0.450
&0.331&0.357\\
&WR-ONE2SET 
&\underline{0.370}&0.378
&\textbf{0.360}&\underline{0.362}
&\underline{0.330}&\underline{0.351}
&\underline{0.428}&\underline{0.452}
&\textbf{0.360}&\underline{0.370}\\
&T5 
&0.336&0.388
&0.288&0.339
&0.302&0.350
& 0.388&0.440
&0.295&0.326\\
&ChatGPT
&0.192&0.158
&0.309&\textbf{0.428}
&0.237&0.189
&0.338&0.258
&0.274&0.252\\
&Ours 
&\textbf{0.387}&\textbf{0.520}
&\textbf{0.359}&\underline{0.360}
&\underline{0.343}&\textbf{0.519}
&\textbf{0.466}&\textbf{0.572}
&\underline{0.336}&\textbf{0.373}\\
\midrule
\multirow{6}{*}{\makecell[cl]{Absent\\Keyphrases}}
&UniKeyphrase
&0.032&0.058
&0.012&0.022
&-&-
&0.026&0.037
&0.022&0.029\\
&PromptKG 
&0.032&0.042
&0.017&0.022 
&-&-
&0.036&0.042
&0.028&0.032\\
&SetTRANS 
& 0.036&0.058
&0.021&0.034
&0.047&0.073
&0.042&0.060
&0.026&0.034\\
&WR-ONE2SET 
&\underline{0.050}&\underline{0.064}
& 0.025&0.034
&\underline{0.057}&\underline{0.074}
&\underline{0.057}&\underline{0.071}
&\underline{0.040}&\underline{0.043}\\
&T5
&0.017&0.034
&0.011&0.020
&0.023&0.043
&0.027&0.051
&0.014&0.020\\
&ChatGPT
&0.025&0.030
&0.014&0.027
&0.002&0.004
&0.003&0.005
&0.002&0.003\\
&Ours 
&\textbf{0.069}&\textbf{0.090}
&\textbf{0.042}&\textbf{0.049}
&\textbf{0.066}&\textbf{0.082}
&\textbf{0.075}&\textbf{0.086}
&\textbf{0.056}&\textbf{0.058}\\
\bottomrule
\end{tabular*}
\label{tab:benchmark}
\begin{tablenotes}
\footnotesize
\item[1] The best results are bold while the second best ones are underlined. $F_1@M$ compares all predictions to ground truths, and $F_1@5$ focuses on the top-5 predictions. When there are fewer than five predictions, we adopt a standard practice in \cite{kp20k, unikeyphrase, one2set}, randomly selecting incorrect keyphrases to fill the vacancies, ensuring a length of five.
\end{tablenotes}
\end{table*}

\subsection{Experimental Setups}
\label{s:exp_settings}
\subsubsection{Datasets}
In our experiments, we use two types of datasets. The first type comprises publicly available benchmark datasets for KG. We select five widely used datasets from previous literature: KP20K \cite{kp20k}, Inspec \cite{inspec}, Krapivin\cite{krapivin}, NUS\cite{nus}, and SemEval\cite{semeval}. Specifically, we train the models on 530,000 of KP20K and spare 20,000 for evaluation. 
In addition, we also perform evaluations on the Inspec, Krapivin, NUS, and SemEval datasets. Specifically, Inspec and KP20K are used to benchmark KG performance in the domain of short scientific documents, while SemEval and NUS focus on long scientific documents. Statistical characteristics of these datasets are contained in the supplemental materials. 

The second type of dataset contains 4,300 patents related to Artificial Intelligence techniques released by the USPTO\footnote{https://www.uspto.gov}. Our dataset shares the same annotation format as \cite{suzgun2024harvard}. We conduct experiments on KG tasks and patent analysis tasks using this dataset. For the KG task, we randomly select 552 patents from the dataset and employ patent experts to annotate the present and absent keyphrases at different levels of each patent, providing ground truths to assess the KG abilities of various methods on patent applications. To validate the effectiveness of KAPPA, we also label the technologies of these patents.

\subsubsection{Baselines}
We use \textit{T5-base} \footnote{https://huggingface.co/google-t5/t5-base} as the PLM backbone of SetPLM\footnote{https://github.com/xxia99/SetPLM} and inherit most of the PyTorch-based settings as \cite{t5}. More implementation details are included in supplemental materials. We compare with the following state-of-the-art baselines: 
\begin{itemize}
\item 
\textbf{UniKeyphrase} \cite{unikeyphrase}: established a unified framework\footnote{https://github.com/thinkwee/UniKeyphrase} for extraction and generation based on a PLM \cite{unilm}.
\item 
\textbf{PromptKG} \cite{promptkp}: utilizes a multi-task framework with keywords for prompt construction for KG. 
\item 
\textbf{T5} \cite{t5}: finetunes a vanilla \textit{T5-base} in One2Seq paradigm with a beam search during inference.
\item 
\textbf{SetTRANS} \cite{one2set}: is the vanilla model incorporating One2Set\footnote{https://github.com/jiacheng-ye/kg\_one2set} and adopts the typical transformer \cite{attention}.
\item 
\textbf{WR-ONE2SET} \cite{wrone2set}: extends SetTRANS to address limitations of $\varnothing$ token over-estimation\footnote{https://github.com/DeepLearnXMU/WR-One2Set}. 
\item
\textbf{GPT-3.5}: is a typical representative LLM released by OpenAI. We use the official prompt \footnote{https://platform.openai.com/examples/default-keywords} for KG
\end{itemize}
 
\subsection{Result Analysis}
\label{s:result}
\subsubsection{Evaluation on Keyphrase Generation Task}

\begin{figure}[!t]
  \centering
  \includegraphics[width=\linewidth]{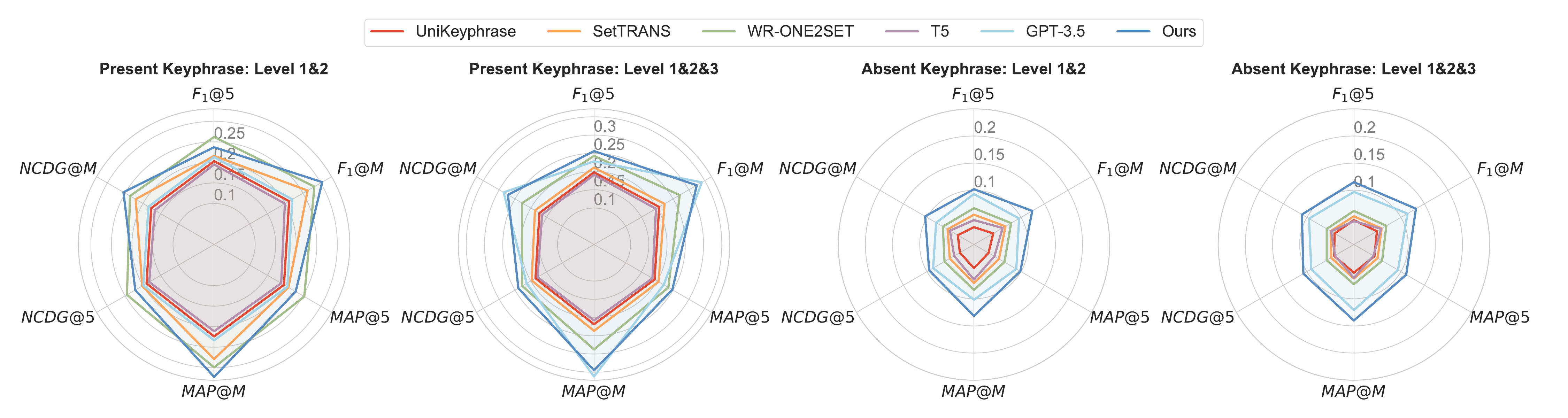}
  \caption{KG Performance on Patent Datasets. MAP@M and NCDG@M compares all predictions to ground truths. MAP@5 and NCDG@5 focuses on the top-5 predictions.}
  \label{fig: patent_kp}
\end{figure}

The first part of experiments aims to evaluate the quality of keyphrases on both benchmark and patent datasets. 
We preprocess data including tokenization, lowercasing, replacing all digits with the symbol \textit{[digit]} and removing duplicated instances as \cite{kp20k, one2set, wrone2set}. To eliminate the influence of morphology, the predicted keyphrases are stemmed by applying Porter Stemmer. We remove all padding keywords from the predictions before comparison. 

\textbf{Performance Evaluation on Benchmark Datasets.} 
Following the common practice \cite{unikeyphrase, keyphrase_survey, one2set}, we concatenate the title and abstract as the source document with a \textit{[EOS]} symbol, and use $F_1$ scores as the criteria of evaluating KG performance. Table~\ref{tab:benchmark} shows that SetPLM outperforms almost all baselines for both present and absent keyphrases. Compared to vanilla SetTRANS, the incorporation of PLM in SetPLM is proved to improve the generative ability with prior knowledge possessed by the PLM.
Compared to vanilla T5 finetuned in One2Seq paradigm, SetPLM demonstrates a potential of One2Set to fortify a PLM's ability on KG task. 
Compared to Unikeyphrase, which also adopts an extractor-generator architecture, SetPLM exhibits significant improvements in KG, further demonstrating the merits of One2Set.
Compared to PromptKG, which also uses keywords as semantic guidance for generation, SetPLM demonstrates superior performance with KCC mechanism, where keywords are directly incorporated into the decoder's input, effectively guiding and constraining KG.

\textbf{Performance Evaluation on Patent Datasets.} 
Figure \ref{fig: patent_kp} illustrates the results of all models' KG performance on real-world patents. In particular, we implement all models to predict keyphrases from two types of subsets of texts from patents. The first type is \textit{Level 1 \& 2}, which contains \textit{Title} and \textit{Abstract} sections. This subset shares a similar length with the benchmark datasets and primarily tests the ability to incorporate domain-specific knowledge when generating keyphrases from patents.
The second type is \textit{Level 1 \& 2 \& 3}, which contains \textit{Title}, \textit{Abstract} and \textit{Claims} sections, examining the ability to generate keyphrases from lengthy documents.
To ensure a comprehensive comparison and mitigate potential bias in our constructed patent dataset, we add two ranking quality metrics: normalized discounted cumulative gain (NDCG) and mean average precision (MAP) as suggested by \cite{mm_keyphrase, kgrl} to evaluate the quality and diversity of the predicted keyphrases.
For absent keyphrases, SetPLM outperforms all baselines, including GPT-3.5, showcasing superior generalization and understanding of the latent information in documents.
Compared to SetTRANS and WR-ONE2SET, SetPLM demonstrates stable KG performance as document lengths increase, thanks to the PHD strategy. This further underscores the robustness and effectiveness of SetPLM in handling long documents. 

\subsubsection{Evaluation on Patent Analysis Tasks}
The second part of our experiments assesses the effectiveness of facilitating patent analysis with portraits. We construct \textit{Pure Portraits} and \textit{Portrait Augmentation} based on keyphrases predicted from sections \textit{Level 1}, \textit{Level 1 \& 2}, and \textit{Level 1 \& 2 \& 3} by different models for comparison. Detailed implementation procedures are included in the supplemental materials.

\begin{table*}
\caption{Patent Classification Performance on Different Levels of Documents. }
\centering
\begin{tabular}{l|l|lr|lr|lr}
\toprule
\multirow{3}{*}{\textbf{Input}}&
\multirow{3}{*}{\textbf{Model}}&
\multicolumn{6}{c}{\textbf{Patent Classification\tnote{1}}}\\
&&
\multicolumn{2}{c|}{\textbf{Level 1}}& 
\multicolumn{2}{c|}{\textbf{Level 1 \& 2}}&
\multicolumn{2}{c}{\textbf{Level 1 \& 2 \& 3}}\\
&&
{Acc.} & {\#Tks.}&
{Acc.} & {\#Tks.}&
{Acc.} & {\#Tks.}
\\
\midrule
Original Text&
-
&0.445&10.37
&0.547&137.31
&0.554&1470.62\\
\midrule
\multirow{4}{*}{\makecell[cl]{Pure Portraits}} &
SetTRANS 
&0.504 & 13.54
&0.480 & 26.09
&0.513 & 30.16\\
& WR-ONE2SET 
&0.531 & 14.18
&0.540 & 31.07
&0.550 & 66.08\\
& ChatGPT
&0.537 & 20.45
&0.557 & 39.55
&0.554 & 90.65\\
& Ours 
&\textbf{0.549}&21.85
&\textbf{0.560}&40.81
&\textbf{0.563}&120.28\\
\midrule
\multirow{4}{*}{\makecell[cl]{Portrait\\Augmentation}}
& SetTRANS 
&0.520 & 23.97
&0.540 & 163.41
&0.553 & 1500.78\\
& WR-ONE2SET 
&0.533 & 24.62
&0.554 & 168.38
&0.555 & 1536.70
\\
& ChatGPT
&0.525 & 30.42
&0.554 & 174.21
&0.556 & 1521.44
\\
& Ours 
&\textbf{0.537}&32.28
&\textbf{0.565}& 178.13
&\textbf{0.573}&1536.70
\\
\bottomrule
\end{tabular}
\label{tab:pat_class}
\begin{tablenotes}
\footnotesize
\item[1] The best results are bold while the second best ones are underlined. \textit{Acc.} refers to accuracy. \textit{Tks.} refers to number of tokens. 
\end{tablenotes}
\end{table*}

\textbf{Patent Classification. }
We select 3,748 patents from our dataset, with CPC labels as ground truths. For a fair comparison, we employ the \textit{bert-base-uncased} model \cite{attention} as a backbone model and perform fine-tuning.
Compared to original texts, Table \ref{tab:pat_class} shows that \textit{Pure Portraits} can achieve comparable performance with fewer tokens. Additionally, \textit{Portrait Augmentation} achieves significant improvements beyond the original texts. This indicates that portraits contain information that is not present in the original texts but can effectively augment the semantic representation of the texts. Notably, our method outperforms the portraits generated by baselines, further demonstrating the quality of the keyphrases generated by our method.
\begin{table*}
\caption{Technology Recognition Performance on Different Levels of Documents. }
\centering
\begin{tabular}{l|l|lr|lr|lr}
\toprule
\multirow{3}{*}{\textbf{Input}}&
\multirow{3}{*}{\textbf{Model}}&
\multicolumn{6}{c}{\textbf{Technology Recognition}}\\
&&
\multicolumn{2}{c|}{\textbf{Level 1}}& 
\multicolumn{2}{c|}{\textbf{Level 1 \& 2}}&
\multicolumn{2}{c}{\textbf{Level 1 \& 2 \& 3}}
\\
&&
{Acc.} & {\#Tks.}&
{Acc.} & {\#Tks.}&
{Acc.} & {\#Tks.}
\\
\midrule
Original Text&
-
&0.631 &9.12
&0.659 &145.87
&0.674 &1757.47\\
\midrule
\multirow{4}{*}{\makecell[cl]{Pure Portraits}} &
SetTRANS 
&0.537 & 4.44
&0.556 & 11.50
&0.630 & 17.38 \\
& WR-ONE2SET 
&0.570 & 4.55
&0.621 & 11.61
&0.648 & 18.87\\
& ChatGPT
&\textbf{0.615} & 9.01
&0.645 & 13.16
&0.667 & 36.60\\
& Ours 
&0.611&9.48
&\textbf{0.648}& 14.79
&\textbf{0.685}&29.76\\
\midrule
\multirow{4}{*}{\makecell[cl]{Portrait\\Augmentation}}
& SetTRANS 
&0.667 & 13.56
&0.667&  157.37
&0.648&  1774.85\\
& WR-ONE2SET 
&0.648 & 13.67
&0.684 & 157.48
&0.704 & 1776.35 \\
& ChatGPT
&0.648 & 18.13
&0.681 & 159.03
&0.743 & 1794.07 \\
& Ours 
&\textbf{0.685}&18.60
&\textbf{0.703}& 160.66
&\textbf{0.759}&1787.23\\
\bottomrule
\end{tabular}
\label{tab:pat_class}
\begin{tablenotes}
\footnotesize
\item[1] The best results are bold while the second best ones are underlined. \textit{Acc.} refers to accuracy. \textit{Tks.} refers to number of tokens. 
\end{tablenotes}
\end{table*}

\textbf{Technology Recognition. }
We select a topic \textit{Techniques related to Language Models} and choose 552 related patents released from 2017 to 2022 from the dataset. Patent experts annotate technologies from these patents, including four major classes: \textit{In-context Learning}, \textit{Reinforcement Learning with Human Feedback}, \textit{Adversarial Networks}, and \textit{Pretraining Strategies}.
Table \ref{tab:pat_class} shows that the portraits can enhance the original texts, but the enhancement is not necessarily determined by the number of tokens contained in the portraits. Instead, the enhancements are strongly related to the quality of the portraits, as represented by the KG performance shown in Table \ref{tab:benchmark} and Figure \ref{fig: patent_kp}.

\begin{table*}[!t]
\caption{Patent Abstract Generation Task based on the Claims and Keyphrases. }
\centering
\begin{tabular}{l|l|ccc|r}
\toprule
\multirow{2}{*}{\textbf{Input}} &
\multirow{2}{*}{\textbf{Model}} &
\multicolumn{3}{c|}{\textbf{Rouge}\tnote{1}} &
\multirow{2}{*}{\textbf{Meteor}} \\
&& \textbf{R@1} & \textbf{R@L} & \textbf{R@LS} & \\
\midrule
\makecell[cl]{Original Text} &
-& 0.377 & 0.242 & 0.231 & 0.249 \\
\midrule
\multirow{4}{*}{\makecell[cl]{Portrait\\Augmentation}}
&SetTRANS & 0.391 & 0.245 & 0.245 & 0.259 \\
&WR-ONE2SET & 0.377 & 0.235 & 0.235 & 0.252 \\
&ChatGPT & 0.401 & 0.252 & 0.252 & 0.260 \\
&Ours & \textbf{0.427} & \textbf{0.241} & \textbf{0.269} & \textbf{0.377} \\
\bottomrule
\end{tabular}
\begin{tablenotes}
\footnotesize
\item[1] The best results are bold. R@M refers to rouge calculated with top-M tokens. R@LS refers to RougeLsum.
\end{tablenotes}
\label{tab:abstract}
\end{table*}
\textbf{Patent Summarization. }
We adopt GLM4 \cite{glm2024chatglm} as the backbone generative model and consider the \textit{Abstract} section as the ground truth. We assess the similarity between the generated abstract and the original abstract using commonly used metrics METEOR and ROUGE \cite{patent_generation, textsum2013}. In Table \ref{tab:abstract}, \textit{Original Input} is the \textit{Claims} section, and \textit{Portrait Augmentation} combines portraits generated by different models with the \textit{Claims}. We can observe that portraits provide additional information to facilitate summarization and effectively augment the original texts in generation.

\begin{figure}[!t]
  \centering
  \includegraphics[scale=0.24]{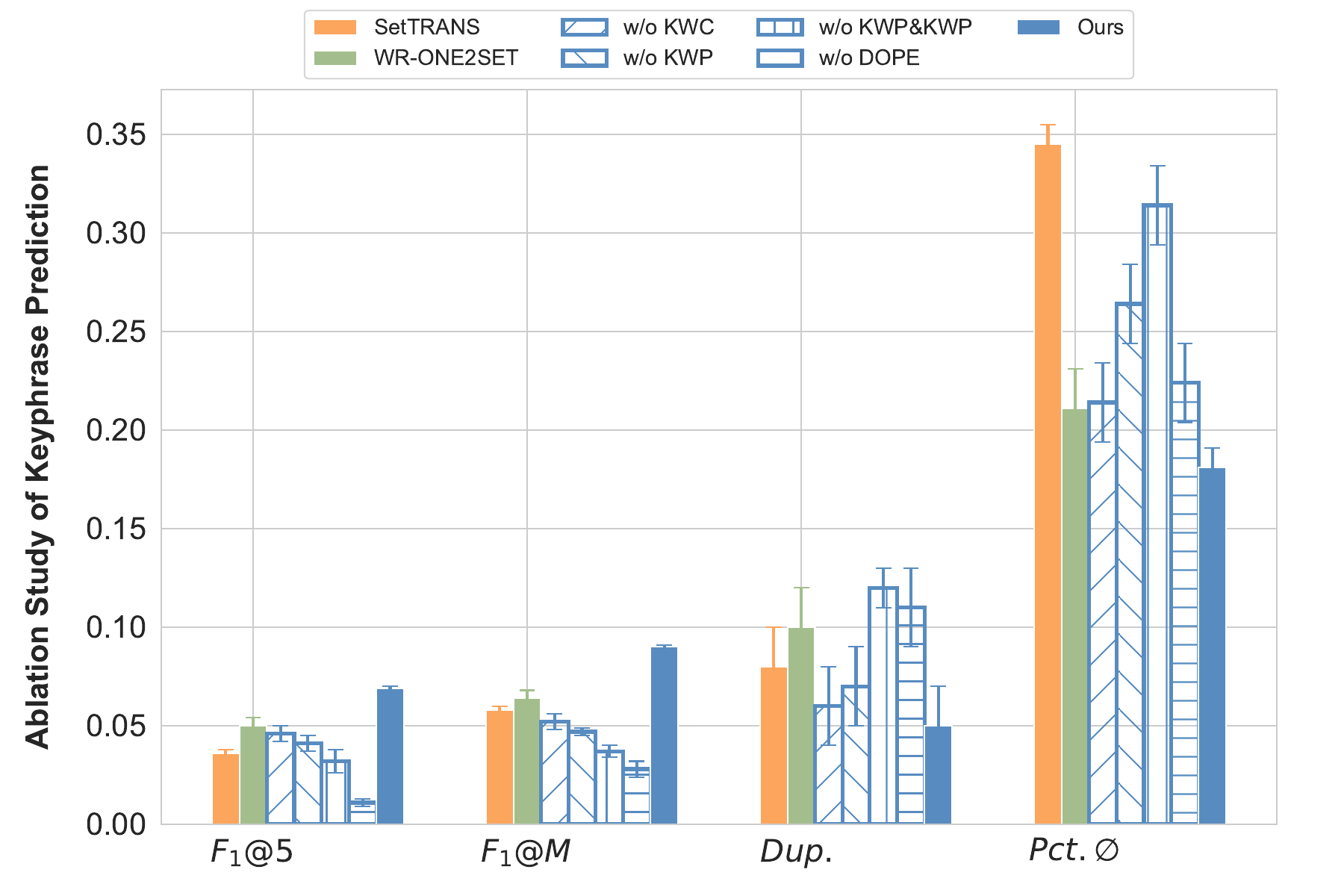}
  \caption{Ablation Study on KP20K Dataset. Dup. refers to the duplication rate of the predicted keyphrases. $Pct.\varnothing$ refers to the ratio of $\varnothing$ tokens contained in predictions.}
  \label{fig: ablation}
\end{figure}

\subsubsection{Analysis of Model Components}
To validate the effectiveness of the components in our model, we conduct an ablation study on the KP20K dataset. Since SetPLM follows a generative One2Set paradigm, where performance is strongly tied to the generation of absent keyphrases, we primarily focus on the $F_1$ scores of absent keyphrases. 

In Figure \ref{fig: ablation}, $w/o (\cdot)$ refers to a variant of SetPLM that excludes component $(\cdot)$.
We can find that SetPLM introduces keywords as decoder input with KCC, leading to substantial decreased duplication ratio of predictions than \textit{w/o KWC} that relies solely on control codes. This observation is further supported by the comparison between \textit{w/o KWP} and \textit{w/o KWC {\rm \&} KWP}.
Compared to \textit{w/o KWP}, SetPLM's keyword-based padding strategy demonstrates notable contributions to alleviated $\varnothing$ token over-estimation. This effect is also observed when comparing \textit{w/o KWC} with \textit{w/o KWC {\rm \&} KWP}. We can also find that KWP introduces diversity to generation which further helps to reduce duplication.
Compared to the variant \textit{w/o DOPE}, SetPLM's keyword-based padding strategy demonstrates notable contributions to alleviated $\varnothing$ token over-estimation. This effect is also observed when comparing \textit{w/o KWC} with \textit{w/o KWC {\rm \&} KWP}. We can also find that KWP introduces diversity to generation which further helps to reduce duplication.

\begin{figure}[!t]
  \centering
  \includegraphics[width=\linewidth]{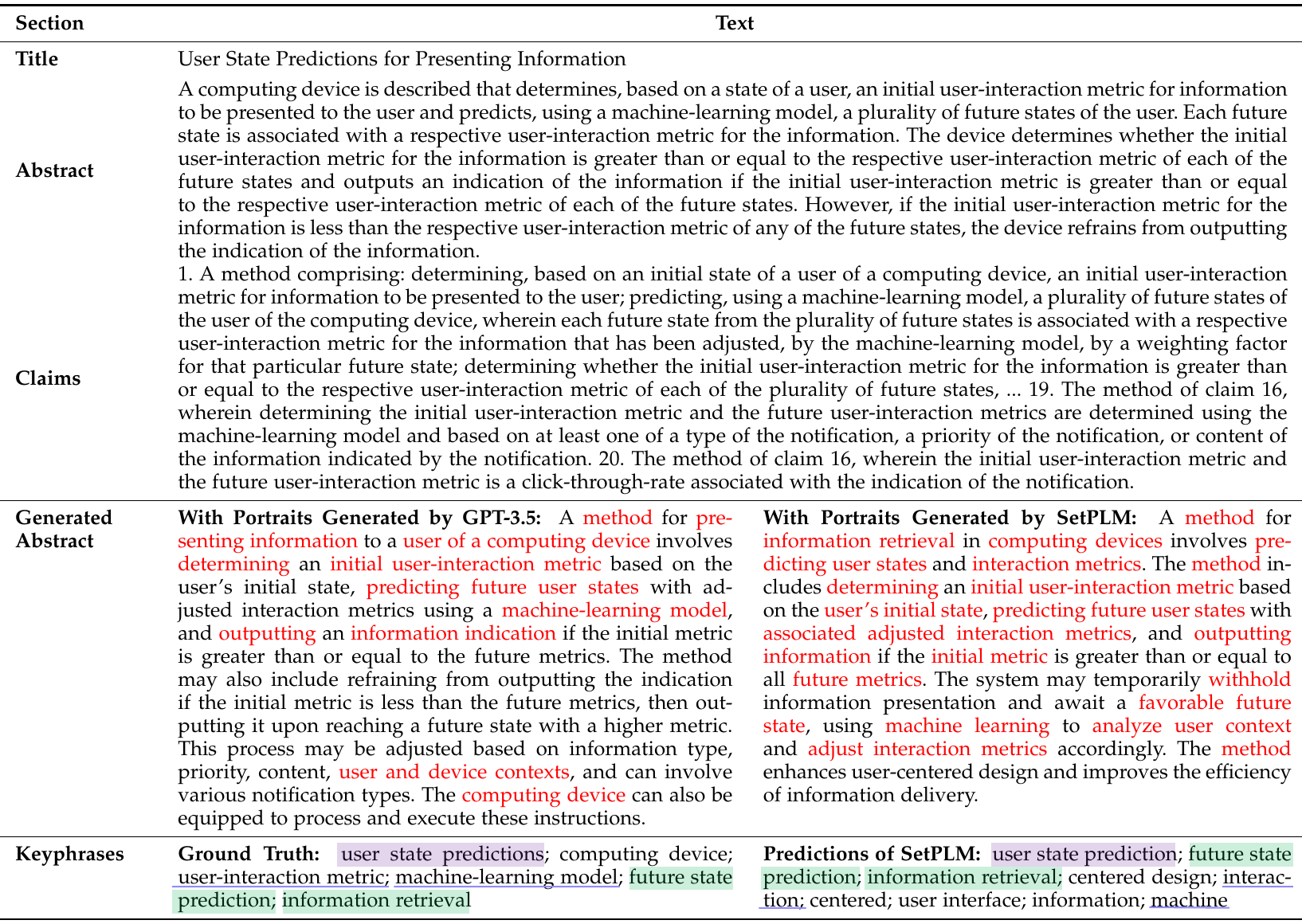}
  \caption{Case Study on a Randomly-selected Patent Released by USPTO.}
  \label{fig: case}
\end{figure}

\subsubsection{Case Study on Real-world Patent Applications}
As illustrated in Figure~\ref{fig: case}, we conduct a case study on a randomly selected patent from the dataset. The section \textit{Generated Abstract} presents the predictions of a backbone generative model with \textit{Portrait Augmentation} constructed by GPT-3.5 and SetPLM, respectively. The texts marked in red in the generated abstracts correspond to the original patent abstract. We observe that the portraits generated by SetPLM result in a more accurate patent summarization, demonstrating superior semantic augmentation.
In the section \textit{Keyphrases}, we include the ground-truth keyphrases labeled by patent experts alongside the predictions made by SetPLM. Matched present keyphrases are highlighted in purple, and matched absent keyphrases are highlighted in green, with partially matched keyphrases underlined. In this case, SetPLM successfully predicts most of the target keyphrases. Notably, SetPLM generates the high-level expression \textit{information retrieval} even though it does not appear in the original document, showcasing its ability to infer abstract concepts.

\section{Conclusion}
\label{s:conclusion}
This paper introduces KAPPA, a framework of keyphrase-based document portraits for patent analysis, which contain domain-specific information within patents, offering representation capabilities with minimal features. We present SetPLM, a model that integrates the One2Set paradigm with PLMs to enhance KG performance on patents. SetPLM addresses the limitations of the traditional One2Set approach, such as duplicated predictions and over-estimation of the $\varnothing$ token. Moving forward, we aim to explore the application of other PLMs, including LLMs, within our framework for patent analysis. Additionally, we seek to develop more efficient and cost-effective training methods for these models.

\end{document}


\title{Supplemental Materials of Paper: KAPPA: A Generic Patent Analysis Framework with Keyphrase-Based Document Portraits}
\maketitle
\section{Dataset Statistics}
\begin{table}
  \caption{Statistics of Benchmark Datasets for Keyphrase Prediction}
  \label{tab:datasets}
  \scalebox{1}{
  \begin{tabular*}{\linewidth}{@{\extracolsep{\fill}}lcccr}
    \toprule
    \textbf{Dataset} & \textbf{$\#$Samples} & \textbf{Avg.$\#$KP} & \textbf{Avg.$\mid$KP$\mid$} & \textbf{Pct.AKP}\\
    \midrule
    KP20k & 20,000 & 5.26 & 2.04 & 37.23\\
    Inspec & 500 & 9.79 & 2.48 & 26.42\\
    Krapivin & 400 & 5.83 & 2.21 & 44.33\\
    NUS & 211 & 10.81 & 2.22 & 45.36\\
    SemEval & 100 & 14.43 & 2.38 & 55.61\\
  \bottomrule
\end{tabular*}
}
\end{table}

\section{Implementation Details}
To implement KAPPA, we use T5-base as the PLM backbone of SetPLM and inherit most of the settings in \cite{t5} based on PyTorch. We employ a AdamW optimizer commonly used in T5 finetuing instead of Adam in conventional SetTRANS and set the learning rate to 0.0003 with a batch size of 36. 

Let $\hat{Y}=\left(\hat{y}_1, \hat{y}_2, \ldots, \hat{y}_m\right)$ and $Y=\left(y_2, y_2, \ldots, y_n\right)$ to be the predicted and target keyphrases, respectively. The common practice is to use only top $k$ predictions with the highest scores for evaluation, where $k$ is a pre-defined constant. In this work, we use $F_1@5$ and $F_1@M$ as criteria. $F_1@M$ compares all predictions to ground truths, and $F_1@5$ focuses on the top-5 predictions. When there are fewer than five predictions, we adopt a standard practice in \cite{kp20k, unikeyphrase, one2set}, randomly selecting incorrect keyphrases to fill the vacancies, ensuring a length of five. To eliminate the influence of morphology, the predicted keyphrases are stemmed by applying Porter Stemmer. We remove all padding keywords from the predictions before comparison.
\subsection{Patent Classification}
In particular, we modify the output dimension of the final fully connected layer of BERT to match the number of classes and utilized full fine-tuning for the model. 
The dataset is split into training and testing sets with an 8:2 ratio. We adopted AdamW as the optimizer and used CrossEntropyLoss as the loss function.
\subsection{Technology Recognition}
We treat technology recognition as a special classification task, where the class labels are defined by technologies strongly related to a practical case interested by the patent experts. 
Similar to the classifiction task, we adopt the "\textit{bert-base-uncased}" model as the backbone, with AdamW as the optimizer and CrossEntropyLoss as the loss function. 
\subsection{Patent Summarization}
Specifically, we use the GLM4 model \cite{glm2024chatglm}. To ensure a fair comparison, the prompt specifies that the generated abstract should have an average length of 130 words.

\section{Additional Experiments and Analysis}
\textbf{Analysis of Keyphrase Quality. }
To investigate the model’s ability to generate diverse keyphrases, we measure the average numbers of unique present and absent keyphrases, and the average duplication ratio of all the predicted keyphrases.

\begin{table}[h]
\centering
\caption{Prediction Diversity on In-Domain and Out-Domain Datasets.}
\begin{threeparttable}
\begin{tabular*}{\linewidth}{@{\extracolsep{\fill}}l|ccc|ccc}
\toprule
\multirow{2}{*}{\textbf{Model}}
& \multicolumn{3}{c|}{\textbf{In-Domain}}
& \multicolumn{3}{c}{\textbf{Out-Domain}}\\ 
&{\#PK}
&{\#AK}
&{\#UK}
&{\#PK}
&{\#AK}
&{\#UK}
\\
\midrule
Oracle\tnote{1}&3.31&1.95& -&3.31&1.95& -
\\
WR-One2Set&4.37&2.43&0.10&4.37&2.43&0.10
\\
SetTRANS &5.01&2.74&0.15&5.01&2.74&0.15
\\
SetPLM  &3.94&2.01&0.05&5.01&2.74&0.15
\\
\bottomrule
\end{tabular*}
\begin{tablenotes}
\footnotesize
\item[1]\textit{ORACLE} refers to the average number of target keyphrases. 
\end{tablenotes}
\end{threeparttable}
\label{tab:diversity}
\end{table}
\newpage
\vfill